# The highD Dataset: A Drone Dataset of Naturalistic Vehicle Trajectories on German Highways for Validation of Highly Automated Driving Systems

Robert Krajewski, Julian Bock, Laurent Kloeker and Lutz Eckstein

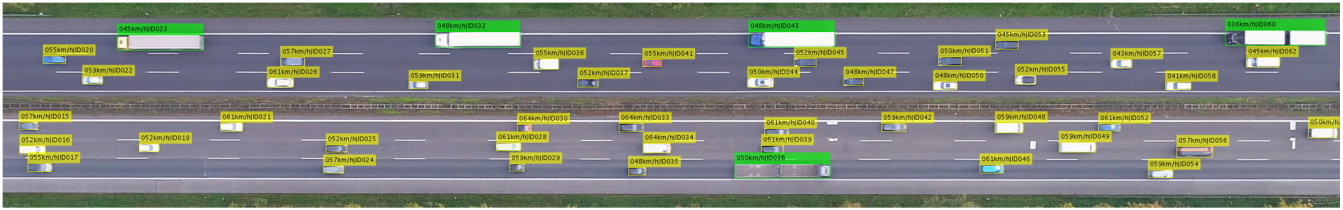

Figure 1. Example of a recorded highway including bounding boxes and labels of detected vehicles. The color of the bounding boxes indicates the class of the detected object (car: yellow, truck: green). Every vehicle is assigned a unique id for tracking and its speed is estimated over time.

*Abstract*— Scenario-based testing for the safety validation of highly automated vehicles is a promising approach that is being examined in research and industry. This approach heavily relies on data from real-world scenarios to derive the necessary scenario information for testing. Measurement data should be collected at a reasonable effort, contain naturalistic behavior of road users and include all data relevant for a description of the identified scenarios in sufficient quality. However, the current measurement methods fail to meet at least one of the requirements. Thus, we propose a novel method to measure data from an aerial perspective for scenario-based validation fulfilling the mentioned requirements. Furthermore, we provide a large-scale naturalistic vehicle trajectory dataset from German highways called highD. We evaluate the data in terms of quantity, variety and contained scenarios. Our dataset consists of 16.5 hours of measurements from six locations with 110 000 vehicles, a total driven distance of 45 000 km and 5600 recorded complete lane changes. The highD dataset is available online at: http://www.highD-dataset.com

## I. INTRODUCTION

A technical proof of concept for highly automated driving (HAD) has already been shown in many demonstrations and test drives. However, existing methods and tools for the safety validation process are not suitable for the complexity of these systems and would be inefficient with regard to costs and time resources [1]. Projects for safety validation and assurance of highly automated vehicles such as PEGASUS [2] and ENABLE-S3 [3] aim to develop a suitable process based on scenarios. A scenario-based approach is also used when conducting impact assessment of automated driving systems [4]. These approaches heavily rely on measurement data from real-world traffic for extracting, describing and analyzing scenarios. In order to cope with the complexity and required level of detail necessary to describe scenarios, five scenario description layers, which are shown in Fig. 2, were defined to structure the description of scenarios on German highways [5] within the PEGASUS project.

Common data sources for safety validation are driving tests, naturalistic driving studies (NDS), field operational tests (FOT) and pilot studies [1]. Test vehicles or series-production vehicles equipped with sensors are used to measure the vehicle´s environment and record the CAN-bus data. A newer approach is the use of infrastructure sensors installed at dedicated roadside masts [6] or at street lights permanently monitoring a certain road segment. However, those measurement methods come with several weaknesses. The necessary quality of the dynamic scenario description and naturalistic behavior of other road users are not always given because of the sensors' physical limitations and the visibility of the sensors.

Thus, we propose to use camera-equipped drones to measure every vehicle's position and movements from an aerial perspective for scenario-based validation. Drones with high-resolution cameras have the advantage of capturing the traffic from a so-called bird's eye view with high longitudinal and lateral accuracy. From this perspective, information about object heights is lost, but vehicles cannot be occluded by other vehicles. However, an object's height has only limited relevance for safety validation and can be estimated from the object type. In altitudes from 100 m up to several hundred meters, the drone is hardly visible from a passing vehicle, which results in completely uninfluenced, naturalistic driving behavior. In our case, a drone was hovering next to German highways and the recordings cover a road segment of about 420 m as displayed in Fig. 3. We use the common term drone throughout the paper for Unmanned Aerial Vehicle, which in our case is a multicopter.

Within this paper, we show the feasibility of this approach and analyze the used methods. Furthermore, we provide a large-scale dataset of naturalistic vehicle trajectories on

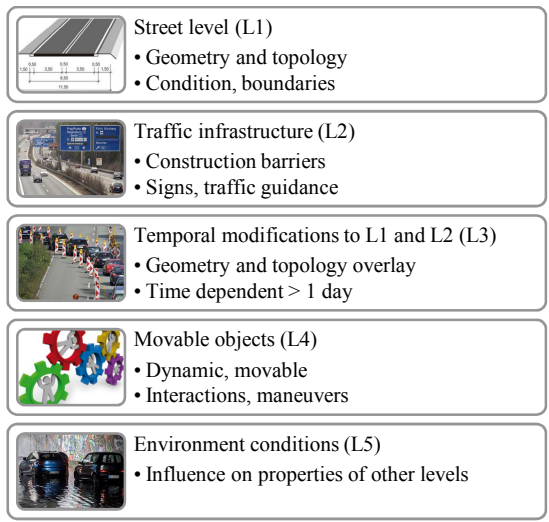

Figure 2. 5-Layer model for the description of scenarios as in [5].

German highways called highD, which stands for highway drone dataset. We compare the highD dataset with other datasets that are used in research. Though the dataset is originally intended for safety validation and impact assessment, we also want to foster research on traffic simulation models, traffic analysis, driver models, road user prediction models and further topics, which rely on naturalistic traffic trajectory data.

## II. PREVIOUS WORK

Within this chapter, we first analyze previous work on the use of drones as sensors for traffic monitoring. Subsequently, we provide an overview on existing datasets for automated driving with focus on the use for safety validation. Because of its closeness to highD, we analyze one of the datasets in detail.

### A. Drones for Recording Road Users

In 2005, the use of video data from camera-equipped drones for traffic monitoring was already examined [7, 8]. However, most of the work had the goal to extract macroscopic data as traffic density, traffic flow and traffic speed [9–12]. As the positions of road users were not extracted with decimeter-accuracy, the resulting trajectories are not suitable for the safety validation of highly automated vehicles.

With the Stanford Drone Dataset [13], a first public dataset with trajectories of multiple road users was created from drone video data. The dataset is intended for the development of pedestrian behavior and interaction models. Recordings were made at eight locations on the Stanford Campus and do not contain public roads. All recordings sum up to a total duration of about 17 hours. Although the dataset contains cars and buses, they account for less than five percent of all road users at seven of the eight locations. At one location, cars account for about 30 percent of the road users, but most of them are parked. Thus, the dataset is not appropriate for safety validation.

To the best of our knowledge, the applicability of aerial video data for safety validation of highly automated vehicles has not been shown yet. Furthermore, currently no public trajectory dataset of vehicle trajectories on highways created with drone video data exists.

### B. Datasets for Automated Driving

There have been several projects dealing with the collection of driving data recorded with onboard sensors within the last ten years [14–17]. In Europe, the project EuroFOT with funding from the European Commission was one of the first large-scale FOTs and ended in 2012. The data of more than 35 million driven kilometers were collected by around 1.200 drivers [17]. The data contains recordings of onboard CAN-bus, raw video, GPS position, front-facing radar and camera. The data is still used in research of impact assessment and safety validation [4]. In the United States, a naturalistic driving study was performed within the second Strategic Highway Research Program (SHRP 2). 3150 volunteers used their vehicles to record 79.7 kilometers between 2010 and 2012. The recordings contain data of front-facing radar, raw-video, vehicle bus data and video of the driver [15]. However, both datasets are not freely available to the public.

In the last few years, several public datasets such as the Next Generation SIMulation (NGSIM) dataset [18, 19], KITTI [20, 21] and Cityscapes [22] were published to foster research on automated driving. Although NGSIM was not originally intended for automated driving but traffic simulation [18, 19], the dataset is now used for automated driving research [23]. As the KITTI and Cityscapes dataset contain single annotated images from vehicle onboard cameras, these datasets are mainly utilized for development of computer vision algorithms such as object detection and scene understanding.

Beside the image set, the KITTI dataset also includes data from laser scanners and object tracks. However, Cityscapes and KITTI mainly focus on urban traffic scenes, while KITTI also contains a few highway traffic scenes. Thus, both datasets have very little relevance for highways scenarios. NGSIM in contrast, focuses on vehicle trajectories on highways and urban traffic roads captured from tall buildings resulting in a bird's eye view. NGSIM is the most similar dataset to the herewith-introduced highD dataset. Thus, we analyze NGSIM in greater detail and compare highD to NGSIM.

### C. NGSIM Dataset

NGSIM is today's largest dataset of naturalistic vehicle trajectories and is widely used for research on traffic flow and driver models [24]. The U.S. Department of Transportation

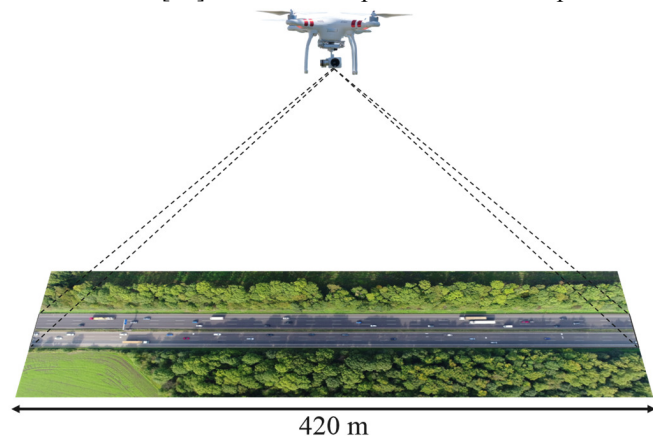

Figure 3. The recording setup includes a drone that hovers next to German highways and captures traffic from a bird's eye view on a road section with a length of about 420 m.

Intelligent Transportation Systems Joint Program Office (JPO) collected video data of traffic in a period from 2005 to 2006. The dataset includes four different recording sites: The Interstate 80 (I-80) in Emeryville, CA, the US Highway 101 (US 101) in Los Angeles, CA, the Lankershim Boulevard (LB) in Los Angeles, CA and the Peachtree Street (PS) in Atlanta, GA. While the highways at the recording sites I-80 and US-101 are comparable to the German Autobahn, the recordings at LB and PS contain urban scenes. Therefore, we only consider I-80 and US-101 in the following. At each site, multiple synchronized video cameras were located on top of an adjacent multistory building recording different overlapping road segments, covering between 500 m and 640 m. The recordings have a total duration of 90 minutes. Next to I-80, seven cameras were installed on top of a multistory building in 97 m height, whereas at US 101 study area, eight cameras were installed on top of an adjacent multistory building with a height of 154 m [25]. Tilted video camera alignments were needed to cover the whole study area.

As shown in previous works [24, 26], raw NGSIM trajectories cannot be used for further analyses. False positive trajectory collisions and physically illogical vehicle speeds and accelerations happen to occur in the dataset. To eliminate erroneous trajectory behavior, [26] refines the longitudinal vehicle movements for a part of the dataset using the trajectories itself. Apart from this, [24] shows that this method is not sufficient for every case and that vehicles first have to be manually re-extracted from the recordings to get improved longitudinal trajectories.

III. ANALYSIS OF MEASUREMENT METHODS FOR SCENARIO-BASED SAFETY VALIDATION

A. Requirements on Measurement Methods

In order to collect data suitable for use in scenario-based safety validation, an appropriate measurement method must be used. In general, the procedure must make it possible to capture all relevant facets of traffic with sufficient accuracy. Although, the specific requirements depend on the desired application, we derived the following five general requirements:

- *Naturalistic behavior:* The behavior of all road users must be naturalistic and uninfluenced by the measurement. Ideally, every road user is unaware of the measurement and thus uninfluenced in its behavior.

- *Static scenario description:* Information belonging to the first three layers of the 5-layer scenario description model [5] including e.g. number of lanes, lane width, speed limits and road curvature must be captured.

- *Dynamic scenario description:* Information belonging to the fourth layer of the 5-layer scenario description model [5] describing the road users' movements must be included in the data. Road users must not be left out because of occlusion but their positions and movements must be measured accurately. Finally, also the data should contain all information regarding the fifth layer of the 5-layer model, which represents environmental conditions.

- *Effort effectiveness:* The total effort consists of the initial effort for setting up the measurement method and permanent effort for operation. Effort effectiveness is the ratio of measured scenarios over combined permanent and initial effort.

- *Flexibility:* Measurements should ideally cover every variance of traffic. Thus, the data should not be limited to a certain road segment but should capture data at every time of the day and every environmental condition.

B. Comparison of Measurement Methods

In the following, we compare the drone-based approach with existing measurement methods in terms of the five requirements. The comparison is displayed as radar chart in Fig. 4. As there exist several measurement campaign setups for the vehicle onboard measurement, which might change in future, we make the following assumptions. First, we consider an NDS setup for vehicles with series-production sensors and we assume that a fused environment model exists only for the front side of the vehicle. Second, for the vehicles equipped with HAD sensors we consider a pilot study and the availability of a 360-degree environment model based on cameras, laser scanners and radar sensors. For the pilot study, we further assume that the driver is permanently aware of the measurements and the test vehicle is recognized as such from the outside by other road users, e.g. due to additionally mounted sensors.

The *Naturalistic behavior* is preserved the best from the aerial perspective as no road user is aware of the measurement. For the NDS study, it can be assumed that the surrounding road users are not aware of the measurement and behave uninfluenced. However, the recorded behavior might not be completely naturalistic since the driver is aware of the recording while driving and the drivers might not truly represent the real driver population. In the pilot study, the behavior of all road users might be influenced as the test vehicles can typically be recognized as such. The untypical appearance of e.g. external-mounted sensors might influence the behavior of the drivers around the measurement vehicle. Roadside infrastructure sensors can generate an accurate overview of traffic in the observed area. However, the sensors are perceived by drivers and might be confused with traffic enforcement cameras leading to atypical driving behavior.

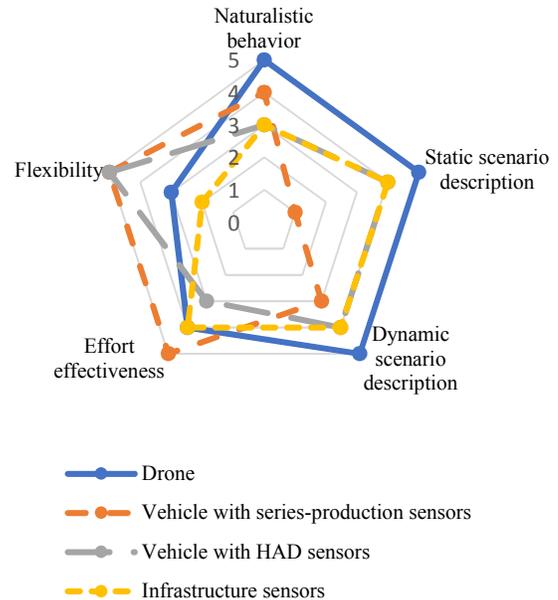

Figure 4. Comparison of measurement methods regarding the use for safety validation of highly automated driving.

The *static scenario description* can be derived from digital map data when infrastructure sensors or drones are used, since both methods are used at a limited number of fixed locations. Additionally, static scenario information can be extracted from the aerial perspective. HAD sensors provide highly accurate localization and comprehensive detections of e.g. lane markings for static scenario description. Finally, NDS might only include inaccurate ego-localization or only simple information, such as the lane markings of the current lane perceived by sensors.

A high quality *dynamic scenario description* can be achieved from the aerial perspective. All vehicles on every lane in both driving directions can be perceived with constant high accuracy. For onboard measurement, the vehicles must be equipped with appropriate sensors for every sensing direction. The data quality of current series-production sensors is typically not sufficient and their data typically hard to access. Furthermore, they can only capture the environment to a comparable limited extent and thus, the scenario cannot be fully described. Vehicles with HAD sensors have 360-degree environmental sensing, but measurement ranges are limited and accuracy is still decreasing with distance. The perception of several sensors must be fused using sensor fusion algorithms. Infrastructure sensors can accurately measure the positions and movements of every object on a certain road segment. However, objects passing by close to the sensors can still occlude other objects.

A NDS has a very high *effort effectiveness* as only minor vehicle modifications might be needed, and the vehicle is operated as if there was no measurement. Road-side infrastructure sensors are also highly effort effective in operation but require a high initial effort for installation. After the initial effort for flight approval, a drone must be operated by an experienced drone pilot. The drone pilot must also drive to the desired measurement location. However, a driven distance of more than 1000 km can be recorded by a drone. The pilot study with HAD sensors comes with high initial effort for setting up the vehicle and selecting the drivers. Typically, the vehicle must be maintained and checked regularly. In the future, the measurement data from vehicles with HAD sensors will become more effort effective when generated unobtrusively with regular series production vehicles.

In comparison, the *flexibility* of measurement vehicles is the highest as they can typically drive on any road and under almost any environmental condition. Drones are basically flexible with regard to the measurement location, but legal flight restrictions and environmental conditions are limiting the measurements to daytime and calm weather conditions. Infrastructure sensors can operate in most environmental conditions and at most locations, but the installation must be approved and coordinated in close cooperation with the road operators. Furthermore, the location cannot easily be changed once installed.

In addition to the advantages and limitations described above, the aerial perspective has drawbacks in terms of online processing. Accurate measurements demand high video resolutions, requiring strong algorithms and unavailable high processing power. However, online processing is not necessary for our purpose to create a dataset. Finally, the aerial perspective has advantages in terms of data privacy protection. Neither the trajectory data nor the raw video data taken by drones are critical in terms of privacy and data protection as no road user can be identified from high altitudes. Data recorded with vehicle onboard sensors are sensitive in terms of privacy and data protection, as private information such as the location or movement patterns can be inferred from the data over time. If cameras are used as infrastructure sensors, they might recognize license plates or even faces. Thus, the raw video data of those cameras are problematic from a data protection point of view.

Summarized, the aerial perspective has several strengths in terms of naturalistic driving behavior, static and dynamic scenario description as well as data privacy protection. Weaknesses are the flexibility compared to vehicle onboard measurement and effort effectiveness compared to onboard measurement with series-production sensors.

## IV. THE HIGHD DATASET COLLECTION PIPELINE

### A. highD at a Glance

The dataset includes post-processed trajectories of 110 000 vehicles including cars and trucks extracted from drone video recordings at German highways around Cologne (see Fig. 5) during 2017 and 2018. At six different locations, 60 recordings were made with an average length of 17 minutes (16.5 h in total) covering a road segment of about 420 m length. Each vehicle is visible for a median duration of 13.6 s. From these recordings, we automatically extracted the vehicles using computer vision algorithms and annotated the infrastructure manually.

The dataset can be downloaded from http://www.highD-dataset.com, while Matlab and Python source code to handle the data, create visualizations and extract maneuvers is provided at https://www.github.com/RobertKrajewski/highD-dataset.

### B. Video Recordings

The videos were recorded in 4K (4096x2160) resolution at 25 fps and were saved at the highest possible quality using the consumer quadcopter DJI Phantom 4 Pro Plus. The drone hovered directly next to German highways to minimize perspective distortions and to record as little of the vehicle sidewalls as possible. The size of a single pixel on the road surface is about 10x10 cm. The recordings only took place during sunny and windless weather from 8 AM to 5 PM to maximize the quality of the recordings and to minimize the need for stabilization caused by movements. Although the quadcopter uses flight stabilization and a gimbal-based camera stabilization, translations and rotations could not be avoided completely. Therefore, videos were stabilized using OpenCV by estimating transformations that map the background in each frame to the background in the first frame of the corresponding recording. Furthermore, the first frame was rotated so that the lane markings are horizontal. Because of these transformations, the actual length of the recorded highway section varies slightly in each frame.

### C. Static and Dynamic Object Annotation

As more than 110 000 vehicles are included in this dataset, a manual annotation was not feasible. Thus, an algorithmic approach was chosen, which is based on state-of-the-art computer vision algorithms. We decided to use an adaptation of the U-Net [27], which is a common neural network architecture for semantic segmentation. The network estimates for every pixel of each frame, if it belongs to a vehicle or to the background. The resulting segmentation map is used to create bounding boxes by detecting pixel clusters belonging to vehicles. Static objects such as lane markings, traffic signs and speed limits were annotated manually, as the effort is negligible in comparison to the annotation of the vehicles.

### D. Track Postprocessing

As the detection runs on each frame independently, a tracking algorithm was necessary to connect detections in consecutive frames to tracks. During this process, detections in two frames were matched by their distances or discarded if no feasible match was found. By doing so, false positive detections could be completely removed. If a vehicle was not detected in few consecutive frames due to e.g. occlusion by a traffic sign, the movements were predicted until a new detection matched the vehicle's track.

Additional postprocessing was applied to retrieve smooth positions, speeds, and accelerations in both x- and y-direction. Using Rauch-Tung-Striebel (RTS) Smoothing [28] and a constant acceleration model, the trajectory of each vehicle was refined taking into account all detections. This improved the positioning error to a pixel size level.

### E. Maneuver Classification

In addition to the raw vehicle trajectories, we have extracted a set of predefined maneuvers for each vehicle to ease work with the dataset, e.g. for analysis. As to our knowledge there is no established list of maneuvers on highways, we use our own list of maneuvers. Each maneuver is detected by a predefined set of rules and thresholds. The maneuvers are not mutually exclusive excepting free driving and car following. We use the definitions of [29] to decide whether a vehicle is influenced by the preceding vehicle or not using a default driver. Critical maneuvers are detected by the rules defined in [30]. The full list of detected maneuvers is:

- *Free Driving (Longitudinal uninfluenced driving):* Driving without being influenced by a preceding vehicle
- *Vehicle Following (Longitudinal influenced driving):* Actively following another vehicle
- *Critical Maneuver:* Low Time to Collision (TTC) or Time Headway (THW) to a preceding vehicle
- *Lane Change:* Crossing lane markings and staying on a new lane

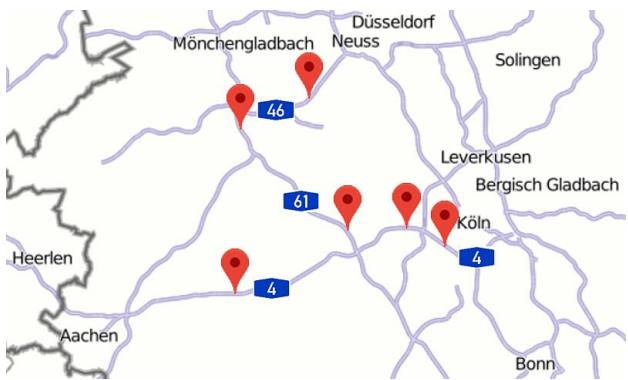

Figure 5. Locations of recordings included in highD. Highways near Cologne were selected by typical traffic density and number of lanes.

Also, the ID, the Distance Headway (DHW), THW and TTC of preceding and following vehicles on the own and adjacent lanes are derived for each vehicle. We provide the scripts for the extraction of these scenarios from the dataset to ease the adjustment of the parameters or the maneuvers.

### F. Dataset Format

The dataset includes four files for each recording: An aerial shot of the specific highway area and three CSV files, containing information about the site, the vehicles and the extracted trajectories. The first file includes the location of the site, driving lanes, traffic signs and speed limits on each lane. A summary of every track including the vehicle dimensions, vehicle class, driving direction and the mean speed is given by the second file. Detailed information like speeds, accelerations, lane positions and a description of surrounding vehicles in every frame are stored for each track in the last file.

## V. DATASET STATISTICS AND EVALUATION

### A. General and Size Comparison of the Datasets

Table I gives a comparison of the amounts of data available in the NGSIM and the highD dataset. While NGSIM provides data of a recording duration of about 90 minutes at two different sites (45 minutes each), highD includes data of more than 16.5 hours of recordings, which were collected at six different sites. In between the recordings, the battery of the drone was exchanged, and the drone was landed/started by the pilot. While highD includes typical German highways with two or three driving lanes per direction, the NGSIM recording sites are highways with five or six driving lanes per direction.

Comparing the number of recorded vehicles, highD contains nearly twelve times as many vehicles as NGSIM. While both datasets contain a negligible amount of motorcycles (as most recordings for highD took place during winter), the ratio between cars and trucks differs. Only 3 % of the vehicles are trucks in NGSIM. This makes it very strongly focused on cars compared to highD with a share of 23 %. While the highD dataset contains a traveled distance that is nine times as high, the total travel time of all vehicles is only almost three times as long, because a lot of dense traffic occurs in the NGSIM dataset.

### B. Variety of Included Data

The highD dataset does not only include more data than NGSIM, but the data have also a higher variety. The main reasons for this are the higher number of recordings and the inclusion of different times of the day and more recording sites. As the histogram of mean track speeds in Fig. 6a shows, highD offers a much broader range of mean speeds. The peaks at 80 km/h and 120 km/h are typical speeds for trucks and cars at the recording sites. Despite an imposed speed limit of 105 km/h at the NGSIM recording sites, tracks with a mean speed above 75 km/h are completely missing. The composition of vehicle types measured by the truck ratio over time varies from 0 % to more than 50% in highD, while it stays below 10 % over time in the NGSIM dataset (see Fig. 6b).

TABLE I.   COMPARISON OF DATA AMOUNTS IN NGSIM AND HIGHD

| Attribute | Dataset | |
|---|---|---|
| | *NGSIM* | *highD* |
| Recording Duration [hours] | 1.5 | 16.5 |
| Lanes (per direction) | 5-6 | 2-3 |
| Recorded Distance [m] | 500-640 | 400-420 |
| Vehicles | 9206 | 110 000 |
| Cars | 8860 | 90 000 |
| Trucks | 278 | 20 000 |
| Driven distance [km] | 5071 | 45 000 |
| Driven time [h] | 174 | 447 |

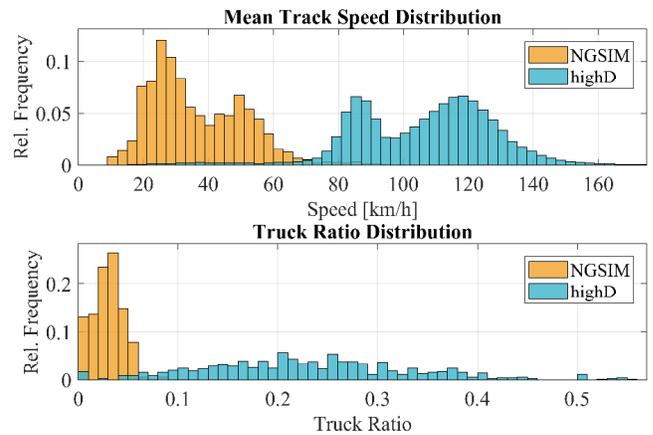

Figure 6. Histogram of a) mean track speeds b) truck ratio over time in NGSIM and highD

### C. Quality Evaluation and Comparison

The initial training set of the semantic segmentation neural network used for the detection consists of about 3000 image patches. The patches include vehicles extracted from ten recordings of different locations with varying light conditions. Augmentation including flipping, adding Gaussian noise and changing the contrast increased the size of the dataset to 12 000 vehicles. The detection thresholds are chosen in favor of a low false negative rate to detect most of the vehicles. Corner cases like unique looking vehicles in the dataset were identified by a strongly changing detected bounding box size in adjacent frames. Afterwards, they were labeled and added to the training set for a second training iteration. Testing on a validation set of images, the trained model detected about 99 % of the vehicles while keeping the false positive detection at 2 %. The resulting mean positional errors of the vehicle midpoint in longitudinal and lateral directions are below 3 cm each in comparison to the manually created labels. The tracking algorithm in the next step removes all false positive detections by simple consistency checks and predicts vehicle locations if vehicles were not correctly detected e.g. due to occlusion.

In comparison to that, an algorithm tracking vehicle fronts was used for the creation of NGSIM. In Fig. 7 the resulting quality for the original NGSIM dataset and highD can be compared. The bounding boxes of NGSIM dataset rarely match the vehicle shapes and several outliers almost exclusively contain the road surface. This matches the analysis in [24, 26], stating that the original results contain numerous errors. These especially occur at transitions between cameras recording different segments of the sites and are caused by the image stitching. Also, due to the necessary rectification and unavoidable occlusions of the tracked vehicle fronts, the tracks have a varying quality. Consequently, unrealistic speeds and accelerations often occur. Also, parallel moving vehicles are sometimes assigned to the same lane resulting in false positive collisions instead of overtaking maneuvers due to errors in the lateral positions. Thus, the original dataset should not be used without preprocessing and [26] released an updated version without infeasible tracks and smoothed longitudinal trajectories. But [24] states, that many errors caused by the tracking system cannot be fixed by filtering alone. Instead, the tracks need to be re-extracted using better algorithms from the non-public original recordings.

Thus, the highD dataset has several advantages due to the use of a single high-resolution camera, a frame rate more than twice as high and a state-of-the-art detection system. In contrast to NGSIM, no further post processing of the tracks is needed, since multiple post processing steps remove all false positive detections and smooth the extracted trajectories.

### D. Maneuver Statistics

Finally, we analyze the occurrences of lane changes and critical maneuvers defined in Section IV. The highD dataset includes more than 11 000 lane changes, of which only 5600 were completely performed in the observed area. This is two times as much as NGSIM includes, while having a lower rate of 0.10 vs 0.45 lane changes per vehicle. One reason for this is that the lower average traffic density and the smaller number of lanes result in fewer lane changes. Also, critical maneuvers occur in the highD dataset. Analyses show that these are mainly caused by tailgating and risky lane change maneuvers.

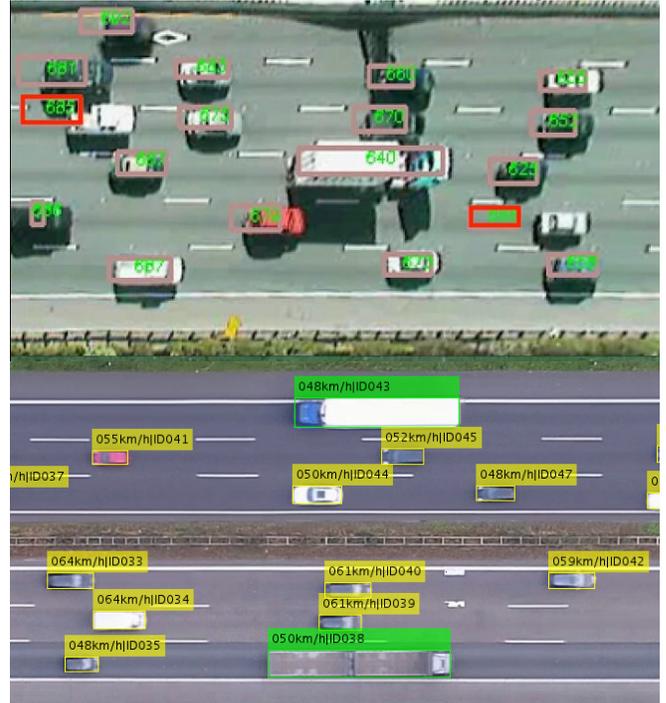

Figure 7. a) NGSIM: Bounding boxes rarely match vehicle shapes. Some bounding boxes almost exclusively contain the road surface (marked red). b) highD: Bounding boxes completely match vehicle shapes.

## VI. ANALYSIS OF EXTRACTED LANE CHANGES

As an example of how the highD dataset can be used for a system-level validation of highly automated driving systems, an analysis of extracted lane change maneuvers was performed. The maneuvers and the surrounding vehicles were parametrized, and statistics were calculated. The frequency distribution of the parameters and parameter combinations can be used as an indication of what kind of lane changes occur under what circumstances. These are necessary statistics for the efficient selection and weighting of test scenarios in simulation or on test tracks.

### A. Lane Change Trajectory Model

Lane changes are typically modeled using sine curves, splines or polynomials [31]. For simplicity, we use a symmetrical model using two separate polynomials for the longitudinal and lateral movement. While choosing a quadratic polynomial for the longitudinal movements, a polynomial of degree five is used for the lateral movement, as the lateral movement is more relevant for lane changes. It is assumed that the vehicle changing the lane has neither a lateral or longitudinal acceleration nor a lateral speed at the beginning and the end of the lane change. Thus, a lane change has five remaining degrees of freedom for which we selected intuitive parameters, which are shown in Fig. 8a. These parameters include the lateral distance to the crossed lane marking and the longitudinal speed at the beginning/end of the lane change. The fifth parameter is the duration of the lane change.

After detecting the lane changes by lane crossings as described in Section IV, the lateral movement determines the beginning and the end of the lane change maneuver. To identify the values for the parameters of the model that describe a trajectory the best, an optimization problem is formulated and solved.

### B. Description of Lane Change Surroundings

The surrounding vehicles of a lane-changing vehicle induce and influence the lane change and thus are included into the statistics. We choose the preceding vehicle on the initial lane and the directly preceding and tailing vehicles on the new lane as the most relevant surrounding vehicles during a lane change. The extracted parameters include the minimal DHW, THW, TTC and the gap size (see Fig. 8b). As shown in [4], these parameters allow an analysis of the inducing conditions and an assessment of the criticality of the performed lane change.

### C. Statistics

As an example of relevant statistics for the validation of highly automated driving systems, we analyze lane changes from the perspective of the tailing vehicle on the new lane. This vehicle is assumed to be automated and perceives the lane change as a cut-in to which it may have to react. From the 5600 parameterized lane changes in highD we extracted 850 cut-in scenarios from the right-hand side. For these, we show the distribution of the THW at that moment the vehicle enters the lane and the linear dependency on ego vehicle speed in Fig. 9.

## VII. CONCLUSION

We propose a new method for collecting data for safety validation of highly automated driving systems and present highD, a new dataset of naturalistic vehicle trajectories on German highways. Using drone captured video data and computer vision algorithms, we automatically extract more than 45 000 km of naturalistic driving behavior from 16.5 h of video recordings. After post processing the vehicle trajectories, a set of four maneuvers and traffic statistics are extracted from the tracks. We demonstrate that highD is appropriate as a data source for safety validation as typical maneuvers and inter- and intra-maneuver probabilities can be extracted. We will publish the dataset upon the release of our paper. Our plan is to increase the size of the dataset and enhance it by additional detected maneuvers for the use in safety validation of highly automated driving.

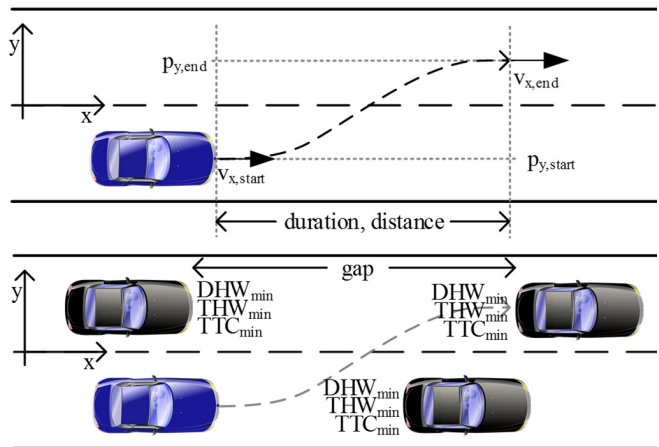

Figure 8. a) Lateral polynomial and extracted parameters of a lane change b) Overview over intracted parameters of surrounding vehicles during a lane change. The blue vehicle symblizes the ego vehicle.

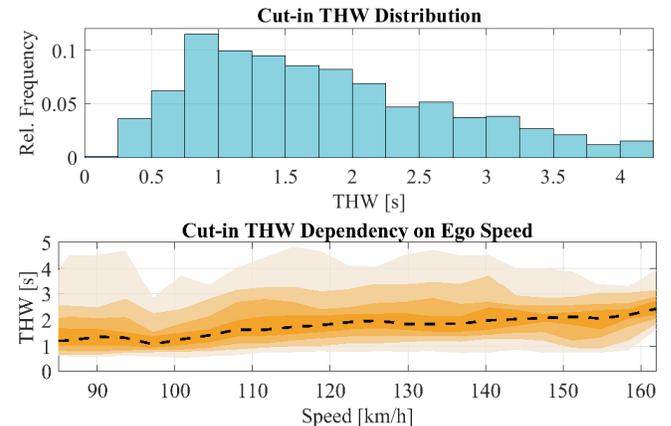

Figure 9. a) Distribution of the tailing vehicle's THW to a lane-changing vehicle at the time it enters the tailing vehicle's lane
b) Dependency of this THW on the tailing vehicle's speed. The dashed line shows the median, while the shades indicate deciles.

REFERENCES

[1] A. Pütz, A. Zlocki, J. Küfen, J. Bock, and L. Eckstein, "Database Approach for the Sign-Off Process of Highly Automated Vehicles," in *25th International Technical Conference on the Enhanced Safety of Vehicles (ESV) National Highway Traffic Safety Administration*, 2017.

[2] W. Wachenfeld, P. Junietz, H. Winner, P. Themann, and A. Pütz, "Safety Assurance Based on an Objective Identification of Scenarios," San Francisco, CA, USA, 2016.

[3] ENABLE-S3, *About the project*. [Online] Available: https://www.enable-s3.eu/about-project/. Accessed on: Mar. 21 2018.

[4] C. Roesener *et al.*, "A Comprehensive Evaluation Approach for Highly Automated Driving," in *25th International Technical Conference on the Enhanced Safety of Vehicles (ESV) National Highway Traffic Safety Administration*, 2017.

[5] G. Bagschick, T. Menzel, and M. Maurer, "Ontology based Scene Creation for the Development of Automated Vehicles," in *29th IEEE Intelligent Vehicles Symposium (IV)*, Changshu, China, 2018.

[6] F. Köster, "Automatisiert und vernetzt im Testfeld Niedersachsen," *DLR magazin*, no. 155, http://elib.dlr.de/117281/1/DLRmagazin-155-DE.pdf, 2017.

[7] A. Puri, "A survey of unmanned aerial vehicles (UAV) for traffic surveillance," University of South Florida, Florida, 2005.

[8] A. Puri, K. P. Valavanis, and M. Kontitsis, "Statistical profile generation for traffic monitoring using real-time UAV based video data," in *Mediterranean Conference on Control & Automation*, Athens, 2007.

[9] M. A. Khan, W. Ectors, T. Bellemans, D. Janssens, and G. Wets, "UAV-Based Traffic Analysis: A Universal Guiding Framework Based on Literature Survey," *Transportation Research Procedia*, vol. 22, pp. 541–550, 2017.

[10] K. Kanistras, G. Martins, M. J. Rutherford, and K. P. Valavanis, "Survey of unmanned aerial vehicles (UAVs) for traffic monitoring," in *Handbook of unmanned aerial vehicles*: Springer, 2015, pp. 2643–2666.

[11] M. A. Khan, W. Ectors, T. Bellemans, D. Janssens, and G. Wets, " Unmanned Aerial Vehicle‐Based Traffic Analysis: Methodological Framework for Automated Multivehicle Trajectory Extraction,*" Transportation Research Record: Journal of the Transportation Research Board*, no. 2626, pp. 25–33, 2017.

[12] B. Coifman, M. McCord, R. G. Mishalani, M. Iswalt, and Y. Ji, "Roadway traffic monitoring from an unmanned aerial vehicle," in *IEE Proceedings-Intelligent Transport Systems*, 2006, pp. 11–20.

[13] A. Robicquet, A. Sadeghian, A. Alahi, and S. Savarese, "Learning social etiquette: Human trajectory understanding in crowded scenes," in *European conference on computer vision*, 2016, pp. 549–565.

[14] R. Eenink, Y. Barnard, M. Baumann, X. Augros, and F. Utesch, "UDRIVE: the European naturalistic driving study," in *Proceedings of Transport Research Arena*, 2014.

[15] K. L. Campbell, "The SHRP 2 naturalistic driving study: Addressing driver performance and behavior in traffic safety," *TR News*, no. 282, 2012.

[16] V. L. Neale, T. A. Dingus, S. G. Klauer, J. Sudweeks, and M. Goodman, "An overview of the 100-car naturalistic study and findings," National Highway Traffic Safety Administration, 2005.

[17] C. Kessler *et al.*, "SP1 D11.3 Final Report," in vol. 3, *EuroFOT Deliverable*, 2012.

[18] J. Colyar and J. Halkias, *NGSIM - US Highway 101 Dataset*. [Online] Available: https://www.fhwa.dot.gov/publications/research/operations/07030/07030.pdf. Accessed on: Mar. 21 2018.

[19] J. Halkias and J. Colyar, *NGSIM - Interstate 80 Freeway Dataset*. [Online] Available: https://www.fhwa.dot.gov/publications/research/operations/06137/06137.pdf. Accessed on: Mar. 21 2018.

[20] A. Geiger, P. Lenz, C. Stiller, and R. Urtasun, "Vision meets robotics: The KITTI dataset," *The International Journal of Robotics Research*, vol. 32, no. 11, pp. 1231–1237, 2013.

[21] A. Geiger, P. Lenz, and R. Urtasun, "Are we ready for autonomous driving? the kitti vision benchmark suite," in *IEEE Conference on Computer Vision and Pattern Recognition (CVPR)*, Providence, RI, USA, 2012, pp. 3354–3361.

[22] M. Cordts *et al.*, "The Cityscapes Dataset for Semantic Urban Scene Understanding," in *29th IEEE Conference on Computer Vision and Pattern Recognition (CVPR)*, Las Vegas, 2016, pp. 3213–3223.

[23] Y. Rahmati and A. Talebpour, "Towards a collaborative connected, automated driving environment: A game theory based decision framework for unprotected left turn maneuvers," in *28th IEEE Intelligent Vehicles Symposium (IV)*, Los Angeles, CA, USA, 2017, pp. 1316–1321.

[24] B. Coifman and L. Li, "A critical evaluation of the Next Generation Simulation (NGSIM) vehicle trajectory dataset," *Transportation Research Part B: Methodological*, vol. 105, pp. 362–377, 2017.

[25] C. Thiemann, M. Treiber, and A. Kesting, "Estimating Acceleration and Lane-Changing Dynamics Based on NGSIM Trajectory Data," *Transportation Research Record: Journal of the Transportation Research Board*, vol. 2088, pp. 90–101, 2008.

[26] M. Montanino and V. Punzo, "Trajectory data reconstruction and simulation-based validation against macroscopic traffic patterns," *Transportation Research Part B: Methodological*, vol. 80, pp. 82–106, 2015.

[27] O. Ronneberger, P. Fischer, and T. Brox, "U-Net: Convolutional Networks for Biomedical Image Segmentation," in *Lecture Notes in Computer Science, Medical Image Computing and Computer-Assisted Intervention – MICCAI 2015*, N. Navab, J. Hornegger, W. M. Wells, and A. F. Frangi, Eds., Cham: Springer International Publishing, 2015, pp. 234–241.

[28] H. Rauch, C. Striebel, and F. Tung, "Maximum likelihood estimates of linear dynamic systems," *AIAA Journal*, vol. 3, no. 8, pp. 1445–1450, 1965.

[29] R. Wiedemann, "Simulation des Straßenverkehrsflusses," Karlsruhe, Germany, 1974.

[30] M. Benmimoun, F. Fahrenkrog, A. Zlocki, and L. Eckstein, "Incident Detection Based on Vehicle CAN-Data within the Large Scale Field Operational Test "euroFOT"," in *22nd International Technical Conference on the Enhanced Safety of Vehicles (ESV)*, Washington, DC, USA, 2011.

[31] W. Yao, H. Zhao, F. Davoine, and H. Zha, "Learning lane change trajectories from on-road driving data," in *IEEE Intelligent Vehicles Symposium (IV)*, Alcal de Henares Madrid, Spain, 2012, pp. 885–890.